\documentclass[letterpaper, 10 pt, conference]{ieeeconf}  

\IEEEoverridecommandlockouts              
\usepackage{graphicx}
\usepackage{amsmath,amssymb,bm}
\usepackage{makecell}
\usepackage{mathtools}
\DeclareMathOperator*{\argmin}{arg\,min}

\newcommand{\bi}{\begin{itemize}}
\newcommand{\ei}{\end{itemize}}
\newcommand{\bd}{\begin{displaymath}}
\newcommand{\ed}{\end{displaymath}}
\newcommand{\be}{\begin{equation}}
\newcommand{\ee}{\end{equation}}
\newcommand{\bem}{\begin{matrix}}
\newcommand{\eem}{\end{matrix}}
\newcommand{\bea}{\begin{eqnarray}}
\newcommand{\eea}{\end{eqnarray}}
\newcommand{\ba}{\begin{array}}
\newcommand{\ea}{\end{array}}
\newcommand{\bc}{\begin{center}}
\newcommand{\ec}{\end{center}}

\title{\LARGE \bf
Interacting humans and robots can improve sensory prediction \\ by adapting their viscoelasticity}

\author{Xiaoxiao Cheng$^1$, Jonathan Eden$^{1,2}$, Bastien Berret$^3$, \\ Atsushi Takagi$^4$ and Etienne Burdet$^1$
\thanks{This work was supported in part by the EC grants PH-CODING (FETOPEN 829186), CONBOTS (ICT 871803) and NIMA (FETOPEN 899626).}
\thanks{$^1$ Department of Bioengineering, Imperial College of Science, Technology and Medicine, SW72AZ London, UK.
{\tt\small \{xiaoxiao.cheng, e.burdet\}@imperial.ac.uk}}%
\thanks{$^2$ Mechanical Engineering Department, the University of Melbourne, Victoria, Australia.}
\thanks{$^3$ CIAMS, Universit\'e Paris-Saclay, Orsay, France; Universit\'e d’Orl\'eans, France; Institut Universitaire de France.}%
\thanks{$^4$ NTT Communication Science Laboratories, 243-0198 Kanagawa, Japan.}%
}

\begin{document}
\maketitle
\thispagestyle{empty}
\pagestyle{empty}

%%%%%%%%%%%%%%%%%%%%%%%%%%%%%%%%%%%%%%%%%%%%%%%%%%%%%%%%%%
\begin{abstract}
To manipulate objects or dance together, humans and robots exchange energy and haptic information. While the exchange of energy in human-robot interaction has been extensively investigated, the underlying exchange of haptic information is not well understood. Here, we develop a computational model of the mechanical and sensory interactions between agents that can tune their viscoelasticity while considering their sensory and motor noise. The resulting stochastic-optimal-information-and-effort (SOIE) controller predicts how the exchange of haptic information and the performance can be improved by adjusting viscoelasticity. This controller was first implemented on a robot-robot experiment with a tracking task which showed its superior performance when compared to either stiff or compliant control. Importantly, the optimal controller also predicts how connected humans alter their muscle activation to improve haptic communication, with differentiated viscoelasticity adjustment to their own sensing noise and haptic perturbations. A human-robot experiment then illustrated the applicability of this optimal control strategy for robots, yielding improved tracking performance and effective haptic communication as the robot adjusted its viscoelasticity according to its own and the user's noise characteristics. The proposed SOIE controller may thus be used to improve haptic communication and collaboration of humans and robots.
\end{abstract}

%%%%%%%%%%%%%%%%%%%%%%%%%%%%%%%%%%%%%%%%%%%%%%%%%%%%%%%%%%%%

\section{Introduction}
Carrying a table together, shared driving, and physical rehabilitation all involve human or robot agents physically interacting to achieve a common goal. In these activities, interaction through direct contact or via an object mediates energy transfer between the agents \cite{Hogan1985ImpedanceI}. Critically, the mechanical connection also enables information transfer about the ongoing movement. Here, humans interacting through the haptic channel have been shown to improve their sensory prediction by exchanging haptic information and then to use it for the integration of sensory information \cite{Takagi2017}. This \textit{haptic communication} is a mechanism of information transfer mediated by the interaction mechanics, in which a more rigid connection yields more direct information transfer at the cost of the partners' freedom of motion \cite{Takagi2018}. 

\begin{figure}[thpb]
\centering
\includegraphics[width=\columnwidth]{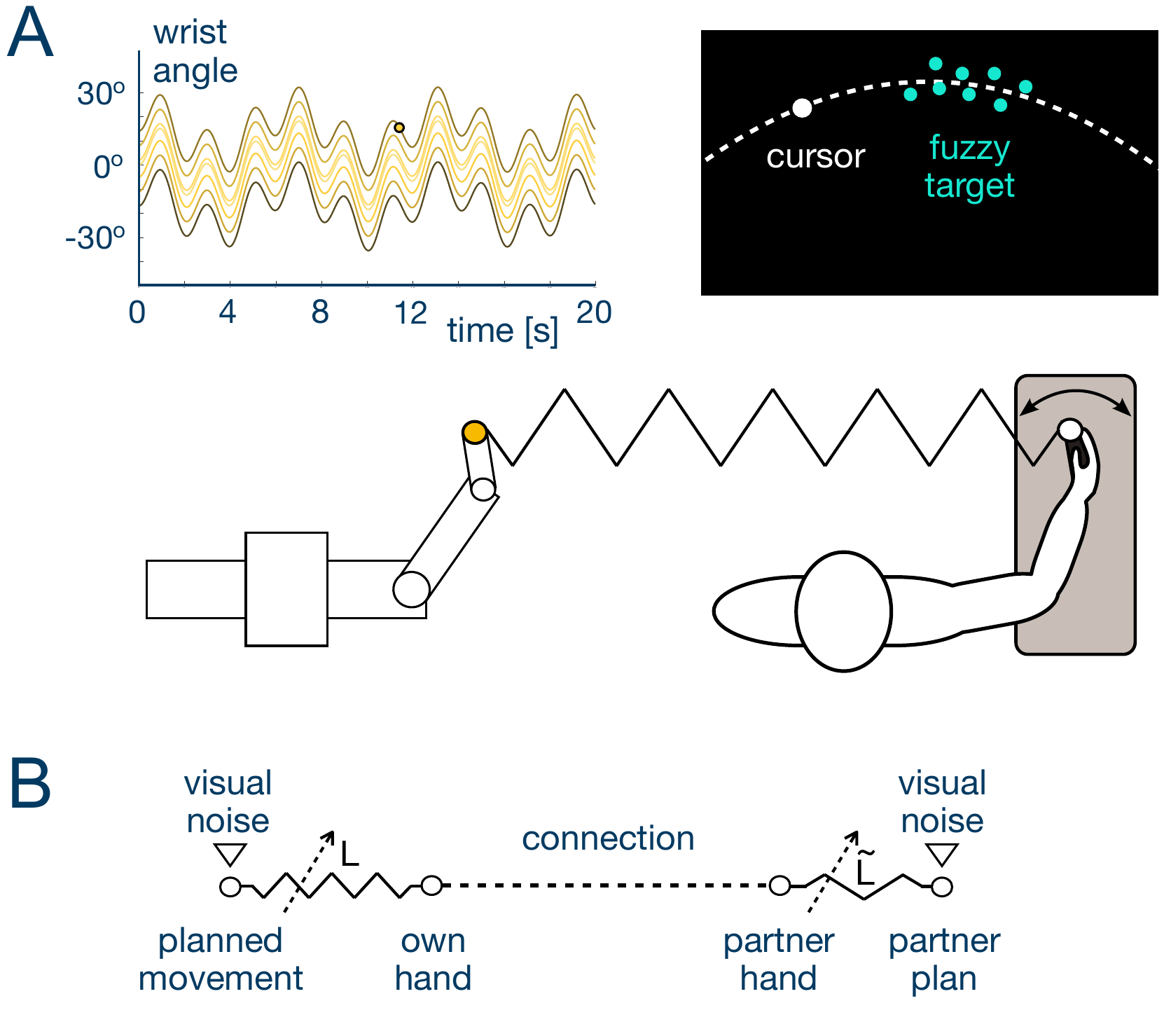}
\includegraphics[width=\columnwidth]{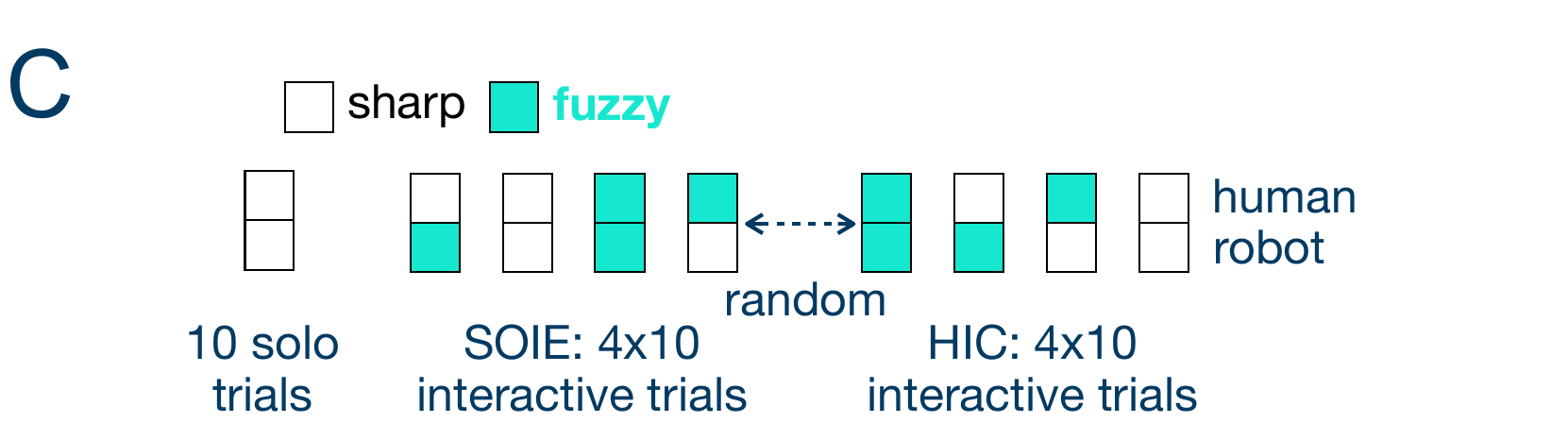}
\caption{Human and robot agents collaborating on a tracking task while exchanging energy and haptic information through an elastic band. (A) Each agent perceives the target movement and their own position, as well as the motion of their partner through the interaction force. The visual and haptic information are perturbed by their respective sensory noise. (B) Each of these soft agents can adapt their viscoelasticity $\{L,\tilde{L}\}$ for following their own planned trajectory to track the target, while considering the partner's movement toward the same target as perceived from the interaction. (C) Protocol of human-robot experiment. Participants started with an initial 10 solo trials to familiarise themselves with the task and interface dynamics, followed by 4 blocks of 10 interaction trials with one controller, and another 4 blocks of 10 trials with the other controller.}
\label{f:interaction}
\end{figure}

Could interacting agents exploit their own mechanics to improve their perception through the shaping of haptic communication? Successful techniques have been developed to regulate the energy transfer between agents  \cite{li2017adaptive, raiola2018development, li2019differential}, but it is unclear how to shape the information transfer. Therefore we first analyze the interaction of two agents considering the stochastic properties of their sensorimotor signals, and how the dynamics depend on their viscoelasticity. Using stochastic nonlinear optimal control theory, we develop a strategy that optimises information exchange and performance in physically connected interacting agents by tuning their viscoelasticity. Through experiments using a common target-tracking protocol (Fig.\,\ref{f:interaction}), we investigate the effect of this viscoelasticity strategy on the interaction between a pair of connected robots, as well as how it can be applied to predict human-human interaction behaviors. Finally, we evaluate whether implementing this technique can enhance human-robot interaction.
% figure
%%%%%%%%%%%%%%%%%%%%%%%%%%%%%%%%%%%%%%%%%%%%%%%%%%%%%%%%%%%%%%%
\section{Stochastic nonlinear optimal control \\ for physical interaction}
We consider two agents such as a robot and its human user that can apply forces on a common object or on each other and tune their viscoelasticity. We assume that these \textit{soft agents} are equipped with (i) position sensing such as proprioception or joint encoders to perceive self-joint positions, and vision, LIDAR or ultrasound to locate objects of interest, and (ii) haptic sensing for the interaction force with a partner or a common object. As a prototype task requiring continuous control, we use the tracking of a visually rendered common target by two agents connected through an elastic band (Fig.\,\ref{f:interaction}A). We consider each agent's planned movement and how their viscoelasticity influences the tracking performance and interaction.

Without loss of generality, we assume that the two agents are a robot physically connected with its human operator. We derive here the robot control law, while the human control can be modelled similarly. The goal of the task is to track a target moving at time $t$ in joint space according to $\eta(t) \in \mathbb{R}^n$. We suppose that, with position sensing noise $\nu \in \mathcal{N}(\delta_\nu,\Sigma_\nu)$, the robot predicts the target motion $\hat{\eta}$ and generates a \textit{motor command} $u$ consisting of a \textit{feedforward part} $\psi (\hat{\eta},\hat{\dot{\eta}},\hat{\ddot{\eta}})$ that has been learned to replicate the robot's dynamics, a \textit{feedback part} $\phi$ to reduce online deviations, and a Gaussian motor noise $\zeta \in \mathcal{N}(0,\Sigma_\zeta)$: 
\be
u(t) \equiv \, \psi \, [\hat{\eta}(t),\hat{\dot{\eta}}(t),\hat{\ddot{\eta}}(t)] + \phi(t) +\zeta(t)
\label{e:control}
\ee
where all these variables are vectors in $\mathbb{R}^n$.

The robot's joint trajectory $q(t) \in \mathbb{R}^n$ is driven by the motor command $u$ and the interaction wrench $\tau$ with the partner. The robot dynamics $\psi$ can thus be represented as 
\be
\psi\left[q(t),\dot{q}(t),\ddot{q}(t)\right] =
u(t) + \tau(t) \,.
\ee
Let these dynamics be linearized around the robot's predicted target $\hat{\eta}(t)$:
\bea 
\label{e:linearisation}
\psi(q,\dot{q},\ddot{q}) \, \cong \,
\psi(\hat{\eta},\hat{\dot{\eta}},\hat{\ddot{\eta}})  + K\xi + D\dot{\xi} + I\ddot{\xi}\,, \\
\xi \equiv q-\hat{\eta} \,, \quad 
\dot{\xi} \equiv \dot{q} - \hat{\dot{\eta}} \,, \quad 
\ddot{\xi} \equiv \ddot{q} - \hat{\ddot{\eta}} \nonumber
\eea
where $K$, $D$ and $I$ are stiffness, viscosity and inertia matrices respectively, which can incorporate dynamics properties of co-manipulated objects. The time variable has been omitted for brevity. We consider that $K, D, I$ change slowly with respect to the position state and assume that they are effectively constant. Using eqs.\,(\ref{e:control}-\ref{e:linearisation}), the robot interaction with the human can be expressed in state-space form 
\bea 
\label{e:state}
\dot{z} \!\!\!\!&=&\!\!\! Az + B(\phi+\tau) + B\, \zeta\,, \quad z \equiv \left[\begin{matrix} \xi\\\dot{\xi} \end{matrix} \right], \nonumber \\
A \!\!\!\!&\equiv&\!\!\!\! \left[\begin{matrix}0_n & 1_n\\-I^{-1}K & -I^{-1}D \end{matrix} \right],\quad  B \equiv \left[\begin{matrix}0_n\\I^{-1} \end{matrix} \right]
\eea
where $1_n$ is the $n$$\times$$n$ identity matrix and $0_n$ is an $n\times n$ matrix formed of 0 elements. 

The feedback control component 
\be
\phi = - L' z\,, \quad z \!=\! \left[\begin{matrix} q - \hat{\eta} \\ \dot{q} - \hat{\dot{\eta}} \end{matrix} \right]
\label{e:vis}
\ee 
uses (the transpose of) the viscoelastic control gains vector $L$ to track the target by reducing the \textit{tracking error} $z$ (Fig.\,\ref{f:interaction}B). Critically, we consider that the robot can change its viscoelasticity to modulate its mechanical response as in an impedance controller \cite{Hogan1985ImpedanceI}. Finally, we assume that the two agents are related by
\be
\tau = - \,\tilde{\tau} \,= \,H \! \left[\begin{matrix} \tilde{q}-q \\ \dot{\tilde{q}}-\dot{q} \end{matrix} \right] = H(\tilde{z}-z+\varepsilon)\,, \,\,\,
\varepsilon \equiv \left[\begin{matrix} \hat{\eta} - \hat{\tilde{\eta}} \\ \hat{\dot{\eta}} - \hat{\dot{\tilde{\eta}}}\end{matrix} \right]
\label{e:tau}
\ee 
where the $\sim$ is used to denote the partner's variables. The $n$$\times$$2n$ matrix $H$ specifies the connection viscoelasticity.

Combining eqs.\,(\ref{e:state}-\ref{e:tau}), the target-tracking closed-loop dynamics yields
\be \label{eq_closedsys}
\dot{z} = \bar{A}z -BL'\nu + BH \mu + B\,\zeta \,,\,\,\, \bar{A} \equiv A \!-\! BH \!-\! BL'
\ee
where $\mu \equiv \tilde{z}+\varepsilon$ is considered as noise conveyed over the haptic channel (\textit{haptic noise} and modelled as a Gaussian biased noise $\mu \in \mathcal{N}(\delta_{\mu},\Sigma_{\mu})$: the bias $\delta_\mu$ represents the partner's tracking inaccuracy, and variance $\Sigma_{\mu}$ considers the degradation of haptic sensation with the connection compliance \cite{Takagi2018, Hendrik2022}. Both $\mu$ and $\zeta$ act as perturbations to the robot tracking dynamics. The haptic noise $\mu$ would be affected by the partner's tracking error and their position sensing noise $\tilde{\nu}$, which can be modelled similarly to eq.\,(\ref{eq_closedsys}).

How should the control stiffness and viscosity gains $L$ be selected? On the one hand, $L$ should reduce the tracking error. On the other hand, coupled with sensory noise, it could amplify the perturbation thus increasing both the error and control effort. To consider this trade-off, we assume that $L$ is adapted to minimize the average error and effort
\be
J \!= \mathbb{E}\!\left[ z'\!(T) Q_T z(T) \!+\! \!\!\int_0^T \!\!\!\! [z'\!(t) Q  z(t) + L'\!(t) R L(t)] \, dt \right]
\label{e:stochOpt}
\ee
In this cost function, $\mathbb{E}[\cdot]$ denotes the expected value, the movement is from time $t=0$ to the finite horizon $t=T$, $R$ is a positive definite matrix and $Q$, $Q_T$ are positive semi-definite matrices. This extends the cost function commonly used to model human motor control \cite{Franklin2008} by considering the stochastics of the sensory signals. 

In eq.\,(\ref{eq_closedsys}) the system states, control variable and noise are interconnected, thus eqs.\,(\ref{eq_closedsys},\ref{e:stochOpt}) yield a nonlinear stochastic optimal control problem which can be solved using the method of \cite{berret2020efficient} assuming that the initial error and all noise in the system are Gaussian distributed. As the tracking error dynamics eq.\,(\ref{eq_closedsys}) is a stochastic differential equation with bilinear drift, the state evolution will be a Gaussian process determined by propagation of its mean $m(t) \!\equiv \mathbb{E}[z(t)]$ and covariance
$P(t) \!\equiv \, \mathbb{E}\{[z(t)\!-\! m(t)][z(t)\!-\!m(t)]'\}$: 
%equation
\bea \label{eq_mean}
\dot{m} \!\!\!&=&\!\!\! \bar{A} \, m -BL'\delta_\nu + BH\delta_{\mu} \nonumber \\
\dot{P} \!\!\!&=&\!\!\! \bar{A} P + P \bar{A}'+BL\Sigma_\nu L' B'+\Omega_{\mu}+\Omega_{\eta} \\
\Omega_{\mu} \!\!\!\!&\equiv&\!\!\!\! BH\Sigma_{\mu}H'B', \quad \Omega_{\eta} \equiv B\Sigma_\zeta B' . \nonumber
\eea
The cost function eq.\,(\ref{e:stochOpt}) can then be represented as
\bea \label{eq_cost_deterministic}
\bar{J} \!\!\!\!&=&\!\!\! m(T)'Q_T\,m(T) + tr(Q_T P_T) + \\ 
&& \!\!\!\!\int_0^T \!\!\!\! \left[\,m'\!(t)\,Q\, m(t) + L'\!RL + tr(QP(t))\right] dt\,. \nonumber
\eea
The stochastic optimal control problem of eqs.\,(\ref{eq_closedsys},\ref{e:stochOpt}) has been converted to the deterministic optimal control problem of eqs.\,(\ref{eq_mean},\ref{eq_cost_deterministic}), where this general formulation uses the mean and covariance of the tracking errors as states. Monte Carlo sampling-based methods \cite{homem2014monte} are used to estimate the mean and covariance of the states from which the impedance value that optimises the cost in eq.\,(\ref{eq_cost_deterministic}) for a certain position sensing and haptic noise condition is determined.

%figure
\begin{figure*}[tb]
%\centering
\includegraphics[width=\textwidth]{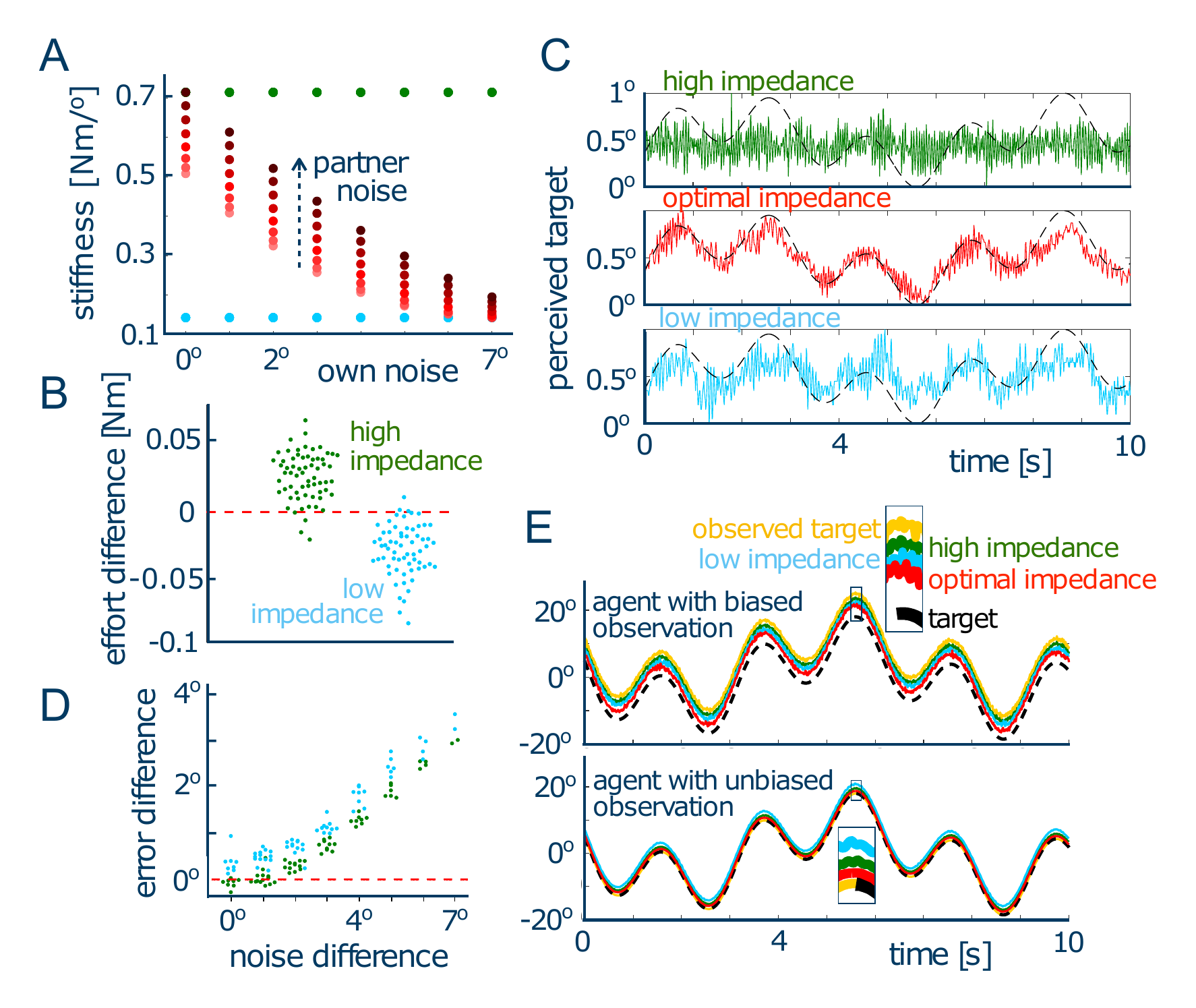}
\caption{Impedance adaptation in robot-robot interaction. (A) How optimal impedance varies with one's own and with partner noise. (B) The effort difference between the SOIE computed impedance and the use of low/high impedance. (C) The perceived target decoded from the interaction torque. With optimal impedance this becomes more similar to the real target trajectory. (D) The tracking error difference. (E) The prediction of the target trajectory by integrating one's own sensing and the interaction with a partner: optimal impedance yields prediction that is superior to both partners with biased and unbiased observations. }
\label{f:rr}
\end{figure*}

The resulting \textit{stochastic-optimal-information-and-effort} (SOIE) controller was implemented on the dual robotic interface of \cite{Melendez-Calderon2011} illustrated in Fig.\,\ref{f:interaction}A. Equipped with two independently actuated wrist flexion/extension handles, as well as torque and angle sensors for both sides, this dual interface enables us to investigate interactions between two human/robot agents in the subsequent sections.

%%%%%%%%%%%%%%%%%%%%%%%%%%%%%%%%%%%%%%%%%%%%%%%%%%%%%%
\section{Impedance adjustment in robot-robot collaboration} \label{section:rr}
To evaluate how the SOIE algorithm affects the information exchange between interacting agents and the corresponding performance, a first experiment was carried out on two robotic interfaces connected by a virtual elastic band of stiffness $17.2$\,Nm/rad. The task consisted of tracking a target moving with trajectory
\begin{eqnarray}
\label{eq:b_traj}
q^*(t) \!\!\!\!&\equiv&\!\!\!\!\!  \, 18.5 \, \sin(\alpha \, t^*) \sin (\beta \, t^*), \\
t^* \!\!\!\!\! &\equiv& \!\!\!\! t + t_0 \, , \,\, 
0 \leq t \leq 10\,\text{s} \, \,\,\, \nonumber 
\end{eqnarray}
using position sensing and the interaction force measured at each robot, where $t^*$ started in each trial from a randomly selected $\{t_0 \in [0,10]\,s \,|\, q^*\!(t_0) \!\equiv\! 0\}$ zero of the multisine function, and the trajectory parameters were set to $\alpha \equiv 2.031\,\text{rad/s}, \,\beta \equiv 1.093\,\text{rad/s}$. Target location sensing was implemented on each robot as the target trajectory subjected to biased Gaussian noise with standard deviation $\sigma_\nu \!=\! 0.05^\circ$ and mean $\delta_\nu \in \{0^\circ,\,1^\circ \dots \,7^\circ\}$. Each robot's impedance control settings, consisting of their stiffness and viscosity, determined the trajectory tracking behaviour. Here for simplicity we assumed that viscosity is proportional to stiffness so that the robot's control impedance can be determined through SOIE using a single parameter as described in the Methods. 

Using the parameters of Table \ref{Sim_parms} in the Methods section, the SOIE could change the robot's control impedance for each noise condition. Fig.\,\ref{f:rr}A shows how the resulting stiffness decreases with one's own (position sensing) noise and increases with the (partner) noise level causing haptic perturbations. The stiffness has smaller change with haptic perturbation compared to the agent's own noise.

We then examined the behaviour induced by the optimal impedance parameters in the 64 noise conditions with each agent having eight sensory noise conditions. One trial was carried out for each of the stiffness values computed with SOIE. Each robot agent used the impedance controller of eq.\,(\ref{e:vis}) with viscosity proportional to the respective stiffness value (with a fixed ratio $0.01$). Controllers with constant minimum or maximum impedance obtained with the SOIE algorithm were also tested for comparison. Fig.\,\ref{f:rr}B shows the root mean square torque with these controllers relative to SOIE as a measure of \textit{effort}. We see that the effort decreases with low impedance ($p<0.001$, paired t-test) %, t-value 11.08) 
and increases with high impedance ($p<0.001$, paired t-test)
%, t-value 12.81) 
relative to the value required with SOIE. 

Meanwhile, we observe in Fig.\,\ref{f:rr}C that the target information exchanged between the partners through interaction torque correlates well (Pearson correlation $r=0.89$, $p<0.001$) with the target movement when using SOIE but not with either high ($r=0.14$, $p<0.001$) or low ($r=0.53$, $p<0.001$) impedance. Correspondingly, SOIE yields higher transmission signal-to-noise ratio ($SNR = 6.99$\,dB) than with the high ($SNR = -2.76$\,dB) and low ($SNR = -0.57$\,dB) impedance. Furthermore, the temporal delay, obtained through the cross-correlations between the predicted and the real target trajectories, indicates that optimal impedance could predict the target with a lower delay ($0.24$\,s) compared to theose generated by fixed high and low impedance controllers ($0.56$\,s and $0.64$\,s respectively). 

These results show that the SOIE controller improves the target information exchange relative to a controller whose impedance is set independently of the noise parameters. As a result of this improved communication, SOIE yields a better prediction with the sensory information than the two constant impedance controllers (Fig.\,\ref{f:rr}E), for both robot agents with biased and unbiased haptic noise. As a consequence, the root-mean square \textit{tracking error} to the target is larger than SOIE with fixed low impedance ($p<0.001$, paired t-test)
%, t-value 10.67) 
and high impedance ($p<0.001$, paired t-test).
%, paired t-Test, t-value 6.87). 
The tracking error difference to the optimal impedance control contribution increases linearly with the noise difference to the partner ($r=0.958$ for the error difference to low impedance control and $r=0.956$ for the difference to high impedance control). These results validate the proposed algorithm and demonstrate the importance of considering both the partner and one's own noise characteristics to select impedance.

%%%%%%%%%%%%%%%%%%%%%%%%%%%%%%%%%%%%%
\section{Muscle adaptations in human-human interaction}
Do humans adapt their impedance to better perceive their partners’ movements when carrying out a collaborative tracking task as suggested by the SOIE model? Previously, we observed in a similar experimental protocol how 22 pairs of physically connected subjects adapted their arm cocontraction while tracking a common randomly moving target using wrist flexion and extension \cite{Hendrik2022}. 
We test here whether the proposed SOIE algorithm could predict the observed human cocontraction adaptation trend. In the experiment, each participant was provided with either sharp or noisy visual feedback of a dynamically moving target. Sharp feedback was with a $8$\,mm large disk, while noisy visual feedback stemmed from a dynamic cloud of eight normally distributed dots. The experiment was performed with four conditions for a dyad of participants: sharp (self) - noisy (partner) (SN), sharp - sharp (SS), noisy - sharp (NS) and noisy - noisy (NN). These four interaction conditions were randomly presented in a block of ten trials per condition.

\begin{figure}[tb]
\centering
\includegraphics[width=\columnwidth]{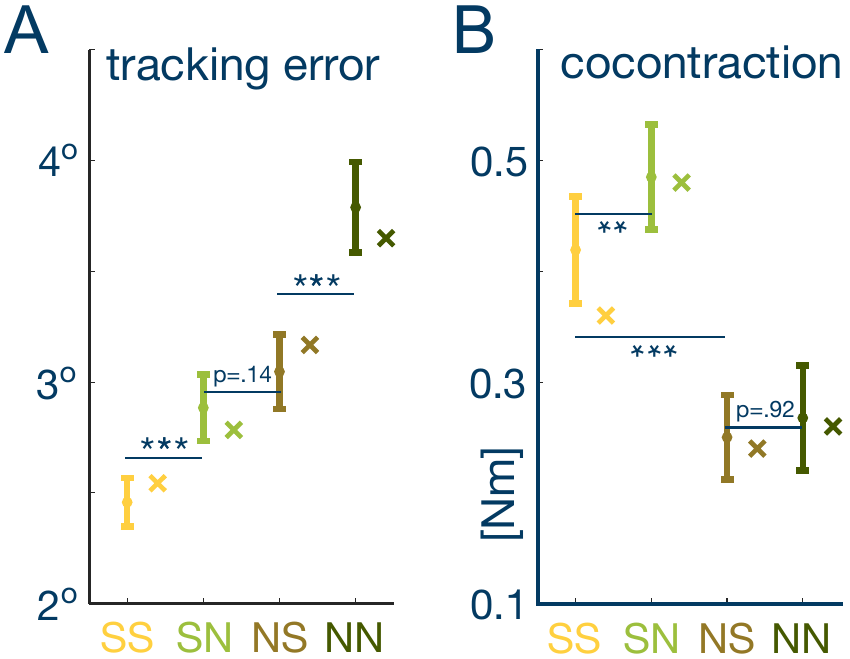}
\caption{Prediction of performance and cocontraction (shown with `x's) in human-human interaction compared to experimental data (shown as intervals defined by the mean value $\pm$ standard deviation) , where SN stands for sharp (self)-noisy (partner), etc. (A) Tracking error with respect to the different conditions. (B) Predicted muscle cocontraction with experimental data. Statistical analysis comparing the differences in experimental data is shown where ** $p<0.01$, *** $p<0.001$.}
\label{f:human-human}
\end{figure}

Figure\,\ref{f:human-human}A shows the wrist cocontraction experimental data. Muscle impedance decreases with an increasing level of their own visual noise and increases with the partner's noise. Furthermore, the participant's own visual noise had a larger effect on cocontraction compared to their partner's noise. This indicates that humans inconspicuously modulate their muscle viscoelasticity to track perceived target movement in the presence of visual blurriness, while simultaneously filtering out haptic noise from mechanical interactions.

To evaluate whether the proposed SOIE algorithm could predict the previously observed human cocontraction adaptation, we compared the experiment data with our new model predicted impedance values in the same sensory conditions. The predicted impedance was found through the optimisation of cost (\ref{eq_cost_deterministic}) subject to the stochastic dynamics (\ref{eq_mean}), while three hyper-parameters: sensory noise in sharp and noisy conditions as well as the effort weight were determined using particle swarm optimisation (PSO) \cite{kennedy1995particle} to minimize the prediction error between the model and experiment data as described in the Methods.

The error could be matched by the model as appears in Fig.\,\ref{f:human-human}A. As can be seen in Fig.\,\ref{f:human-human}B, the cocontraction was then modulated by both their own and the partner’s visual noise as predicted by the SOIE model. Consistent with experimental data, cocontraction was largely influenced by their own visual noise, and weakly affected by partner’s visual noise through the haptic connection, while the tracking error was influenced by both sources of noise. This indicates that the SOIE model can predict the muscle cocontraction adaptation in interacting humans and captures the interactive tracking dynamics, a testament of its predictive power.
%figure

%%%%%%%%%%%%%%%%%%%%%%%%%%%%%%%%%%%%%%%%%%%%
\section{Robot impedance tuning to improve human collaboration}
If a collaborative robot adjusts its viscoelasticity using the SOIE algorithm, could this improve collaboration with its user as was observed in humans? To test this, we implemented a human-robot interaction task similar to the tracking experiment of previous sections. A handle of the Hi5 robotic interface, controlled using an impedance computed with the SOIE algorithm, was connected to the other handle operated by a human participant via a virtual elastic band. Both the human and robot had to track a target moving relatively fast based on eq. (\ref{eq:b_traj}) with $\alpha=3.04\,\text{rad/s}$ and $\beta=2.51\,\text{rad/s}$. 

The trials were carried out with four different noise conditions where the human or the robot had either sharp or noisy sensory condition: sharp (robot) - noisy (human) (SN), sharp - sharp (SS), noisy - sharp (NS) and noisy - noisy (NN). The same visual feedback of either sharp or noisy condition was implemented for the human as in the previous human-human experiment \cite{Hendrik2022}. In the \textit{sharp robot} condition, the robot controller received the precise target trajectory. In the \textit{noisy robot} condition, a biased Gaussian noise was introduced relative to the reference target trajectory, with a mean of $\delta = 7.01^\circ $ and a standard deviation of $\sigma = 0.05^\circ$. Participants were informed of the overall tracking error at the end of each trial to facilitate their adaptations to the noise conditions. 

Considering both sources of noise, the SOIE algorithm computed the optimal impedance through eq.\,(\ref{eq_cost_deterministic}). This was compared with constant high impedance control (HIC) typically used as trajectory guidance \cite{ivanova2020motion}, 
as shown in Fig.\,\ref{f:rr}A. The experimental protocol is depicted in Fig.\,\ref{f:interaction}C. Participants started with an initial 10 solo trials to familiarise themselves with the task and interface dynamics, followed by 4 blocks of 10 interaction trials with one controller, and another 4 blocks of 10 trials with the other controller. The order of the two controllers was randomized within the participant group, where the 4 blocks of each controller used the 4 randomized noise conditions. 

\begin{figure*}[!bt]
\centering
\includegraphics[width=1\textwidth]{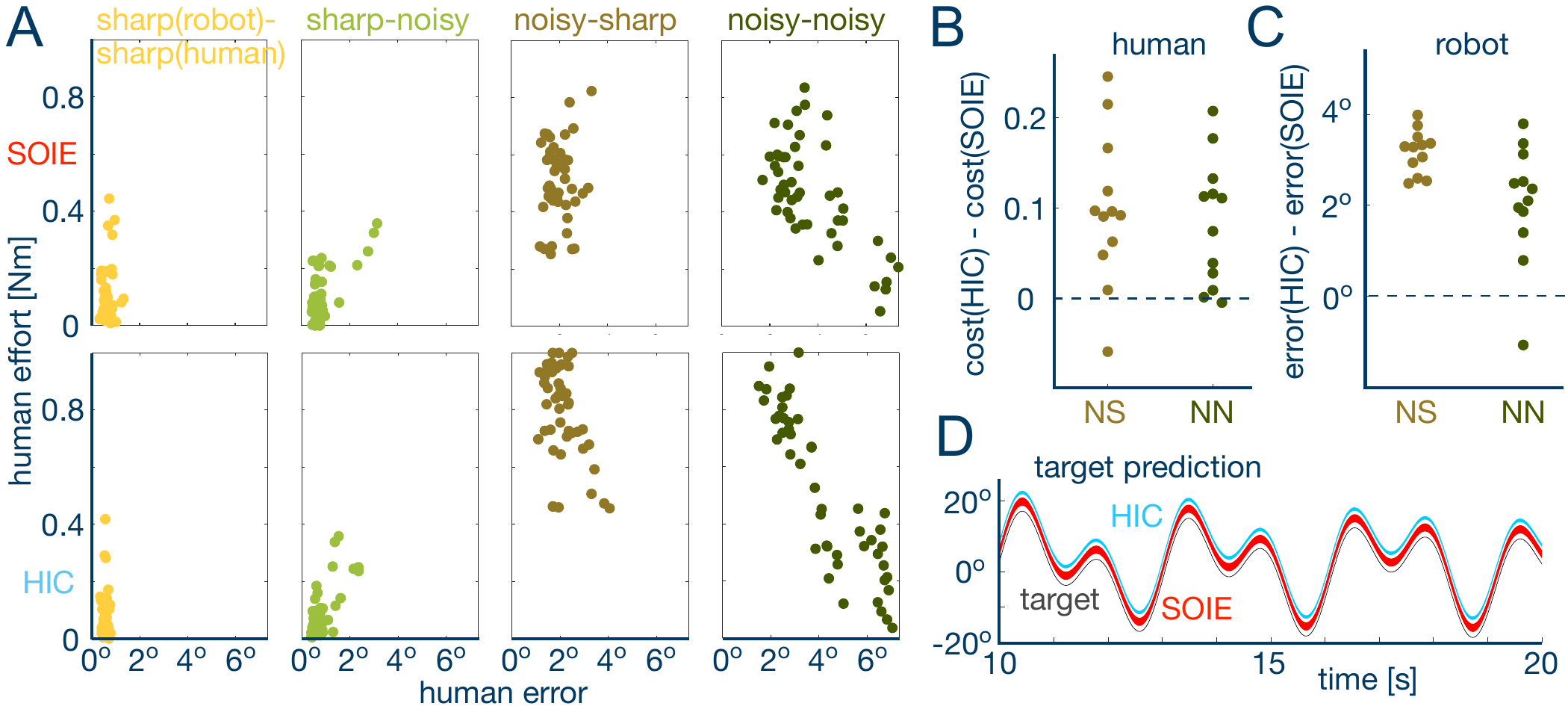}
\caption{Tracking results for the human-robot collaboration experiment. (A) The 12 participants’ tracking errors and effort in the last four trials of all the conditions. The differences in the average normalised costs (B) and tracking error (C) of the robot agent for the SOIE controller and high impedance control (HIC) when the robot had noisy sensing conditions. (D) The predicted target movement. This is more accurate with the SOIE controller compared to the HIC when integrating the interaction torque from human partners. The width of each strip represents the standard deviation of the target prediction among the 12 participants. }
\label{f:hrperformance}
\end{figure*}

When the robot had sharp sensory information, the SOIE used either the same high control impedance ($40.93$\,Nm/rad in the SN condition) as the HIC or another relatively high impedance ($29.12$\,Nm/rad for SS). Therefore the tracking error and effort of 12 participants in their last four trials of each sensory condition are similarly small with SOIE and HIC (Fig.\,\ref{f:hrperformance}A, $p>0.05$ for either error or effort, paired Wilcoxon test).

When the robot's sensing becomes noisy, the SOIE reduces its stiffness to $8.74$\,Nm/rad for NS and $11.20$\,Nm/rad for NN. In the NS condition, this reduced the human effort ($p<0.05$, paired Wilcoxon test) relative to the HIC condition where the human had to correct for the robot's error, without clearly affecting the tracking error ($p>0.05$). In the NN condition, participants using the HIC either exerted a large effort to achieve a small tracking error or spent a small amount of effort resulting in large tracking errors. Using the SOIE, most participants could compromise between these two strategies, resulting in a lower average of both error and effort. 

Compared to the HIC, using the SOIE reduced the effort plus error cost of \cite{Franklin2008} both in the NS ($p=0.024$, paired Wilcoxon test) and NN ($p = 0.019$) conditions (Fig.\,\ref{f:hrperformance}B). Fig.\,\ref{f:hrperformance}C shows that using the SOIE significantly improved the tracking performance of the robot, with an average reduction of $2.08^\circ$ in NS and $1.52^\circ$ in NN (both with $p<0.001$, paired Wilcoxon test). This arose as the SOIE predicted the target movement more accurately than the HIC, as illustrated for the NS condition in Fig.\,\ref{f:hrperformance}D.

%figure

%%%%%%%%%%%%%%%%%%%%%%%%%%
\section{Discussion}
The ease with which humans carry out common actions such as moving furniture or dancing tango may rely on their abilities to understand each other's motion \cite{Takagi2017, sawers2014perspectives} and to modulate the interaction through selective muscle activation \cite{Hendrik2022}. To develop such abilities in robots and enable them to better assist human users, we considered robots and humans as agents able to regulate their viscoelasticity, and analyzed the intermingled dynamics of their mechanical properties and stochastic sensory signals during their physical interaction. Using nonlinear stochastic optimal control we then determined how these agents could modulate their mechanical impedance to improve performance with minimal effort. 

The resulting stochastic-optimal-information-and-effort controller (SOIE) was first validated through robot-robot experiments. It was then shown to result in improved human-robot interaction through the modulation of the robot's viscoelasticity according to its sensory noise and human tracking accuracy. Moreover, the predicted body adaptation of two interacting humans closely corresponded to experimental observations \cite{Hendrik2022}. This result suggests that the SOIE is a suitable model of human impedance adaptation, and can be used to improve human-robot interaction. It is noted that the method is currently designed for constant impedance within a trial such that Monte Carlo sampling methods could serve as a simple method to solve the stochastic optimisation problem. This limitation could be addressed in future by integrating a numeric optimisation solver as in \cite{berret2020stochastic}  to enable continuous impedance adjustment in a dynamic environment.

Previous human motor learning and adaptation models \cite{Franklin2008, theodorou2010generalized, Herzfeld2014, Takiyama2015, Li2018ieeeTRO, berret2020stochastic} only considered the interaction of an arm with a force field environment. Recent controllers for multi-agent physical interaction that consider energy exchange through the haptic channel \cite{raiola2018development, li2019differential} neglected the associated information transfer. SOIE is (to our knowledge) the first controller that considers the interaction with a reactive partner and the stochastics of information transfer. As a consequence, it can explain the observed adaptation of mechanical impedance found in interacting humans \cite{Hendrik2022}, which is not predicted by previous models such as \cite{Franklin2008, theodorou2010generalized, Li2018ieeeTRO, berret2020stochastic}.

Robots are increasingly used in physical contact with humans, for physical rehabilitation, industrial collaborative robots, teleoperated surgical robots, or semi-autonomous cars offering shared control through the steering wheel. With enhanced ability to perceive the environment and plan their motion, these contact robots may adapt their impedance to track their planned movement according to their sensory information and human user's actions using SOIE, with performance and behaviour benefits suggested from the results of the presented experiments. Interestingly, the haptic communication mechanism between humans revealed in \cite{Takagi2017} and extended in this paper is similar to the adaptive and reactive robot controller tested in our experiment. Thus we can envision collectives of humans and robots working together in optimal collaboration using the SOIE mechanism, which may offer similar benefits to those observed in human collectives \cite{Takagi2019}.

%%%%%%%%%%%%%%%%%%%%%%%%%%
\section{Materials and Methods}
\subsection{Optimal impedance computation with Monte-Carlo sampling}
A human agent or a robot can modulate their mechanical impedance by changing their muscle cocontraction or control impedance in eq.\,(\ref{e:vis}) according to $L  \equiv\lambda \, [1\, \text{Nm/}^\circ,\,0.01\, \text{Nms/}^\circ]'$ using the \textit{impedance parameter} $\lambda$. To obtain the optimal impedance with $\lambda^*$, eq.\,(\ref{eq_cost_deterministic}) is used to convert the cost function into deterministic values with the evolution of the mean and covariance eq.\,(\ref{eq_mean}). The agent's own sensory noise $\nu$, haptic noise $\mu$ are both stochastic variables. A Monte Carlo simulation with 500 trials was conducted to estimate the mean and covariance: $\hat{m}(t) = z_i(t)/500$, $\hat{P}(t) = e_i(t)e_i(t)'/500$ where $e_i(t) = z_i(t) - \hat{m}(t)$ and $i = 1,2,...,500$. Optimal mechanical impedance was then determined from
\be
\lambda^* = \displaystyle \argmin_{0\leq\lambda\leq 1} \bar{J}(\hat{m},\hat{P})
\ee
using the parameters in Table I, where the equivalent deterministic cost eq.\,(\ref{eq_cost_deterministic}) is calculated through the Euler integration method with a 0.01\,s time step: 
\[\bar{J}(\hat{m},\hat{P})=  \sum_{i=1}^{2000}\hat{m}_i'\,Q\,\hat{m}_i + \lambda'_i\,R\,\lambda_i + tr(Q\hat{P}_i)\,.\]

\subsection{Hi5 dual robotic interface}
The experiments were carried out using the dual wrist robotic interface Hi5 \cite{Melendez-Calderon2011} sketched in Fig.\,\ref{f:interaction}A. Each handle of the Hi5 interface is connected to a current-controlled DC motor (MSS8, Mavilor) that can exert a maximal torque $15.38$\,Nm. The wrist angle is measured with a differential encoder (RI 58-O, Hengstler) and the exerted torque in the range of magnitudes [$0$,$11.29$]\,Nm with a (TRT-100, Transducer Technologies) sensor. The two handles are controlled at 1\,kHz using Labview Real-Time version 14.0 (National Instruments) and a data acquisition board (DAQ-PCI-6221, National Instruments), while the data was recorded at $100$\,Hz.

%%%%%%%%%%%%%%%%%%%%%% table
\begin{table}[tb]
\caption{Parameters used in simulations}
\label{Sim_parms}
\begin{center}
\begin{tabular}{cll}
%\Xhline{1\arrayrulewidth}
\hline \textbf{Symbol}& \textbf  {Definitions} &\textbf{Value}\\  \hline
 $T$ & task duration & 20 s
\\ \hline
$dt$ & step & 0.05\,s
  % \\ \hline
  % $\xi_0$ &initial error & [0.01\,rad,\ 0\,rad/s]
  \\ \hline
  $I$ & inertia of interface with wrist  &  0.0080\,kg$\cdot$rad$^2$
  \\ \hline
  $H$ & connection stiffness &   $17.32$\,Nm/rad
    \\ \hline  
  $\delta_{\nu1}$ & mean of sharp visual noise  &  $2.56^\circ$ 
  \\ \hline
$\delta_{\nu2}$ & mean of fuzzy visual noise   &  $3.67^\circ$ 
  \\ \hline  
  % $\delta_{\mu1}$ & haptic noise mean 1  &  $6.62^\circ$ 
  % \\ \hline  
  % $\delta_{\mu2}$ & haptic noise mean 2 &  $8.63^\circ$   
  % \\ \hline
$\sigma_{\nu},\sigma_{\mu}$ & noise variance  &  $0.05^\circ$  
  \\ \hline
  % $b$ & wrist viscosity &  0.064\,kg\,rad$^2$/s
  % \\ \hline
  % $k$ & wrist stiffness & 1.28\,kg\, rad$^2$/s$^2$  
  % \\ \hline
  % $\Sigma_\nu$ & visual noise cov. &    $\left[\begin{matrix} 0.018 \!\!\!\!&\!\!\!\! 0\\ 0 \!\!\!\!&\!\!\!\! 0.00018 \end{matrix}\right]$ rad$^2$
  % \\ \hline
  % $\Sigma_\zeta$ & trajectory noise cov. &    $\left[\begin{matrix} 0.035 \!\!\!\!&\!\!\!\! 0\\ 0 \!\!\!\!&\!\!\!\! 0.00035 \end{matrix}\right]$ rad$^2$
  %   \\ \hline
  % $[\alpha,\beta]$ & human control vector &  [2,\ 0.2]\,Nm/rad
  %   \\ \hline
  $Q$ & error weight matrix &   $\left[\begin{matrix} 1/\text{rad}^2 \!\!\!&\!\!\!\!\, 0\\ 0 \!\!\!&\!\!\!\!\, 0.01\text{s}^2/\text{rad}^2
  \end{matrix}\right] $
    \\ \hline
  $R$ & effort weight &  4.02\,\text{/s}$^2$
\\ \hline
  $\rho$ & viscoelasticity ratio &  0.01\text{s}
 \\ \hline
%     \\ \hline
%   $\alpha$ & conversion ratio &  0.013  & -  
% \\ \Xhline{1\arrayrulewidth}
\end{tabular}
\end{center}
\end{table}

\subsection{Robot-robot experiment}
In the robot-robot experiment, each of these agents had two noise conditions. In the \textit{sharp condition} the target reference trajectory was directly tracked by the robot, while in the \textit{noisy conditions} seven biased target positions $\delta_\nu$ in the range of $[1^\circ,7^\circ]$ were passed to the feedback controller for trajectory tracking
\begin{equation}
\phi = -L'z\,,\quad z = \left[\begin{matrix} q-\hat{\eta} \\ \dot{q}-\dot{\hat{\eta}} \end{matrix} \right], \quad \hat{\eta} = \eta + \delta_\nu
\end{equation}
where the impedance parameters $L$ were computed offline using SOIE as described above. The highest and lowest impedance parameters found by SOIE were used for comparison and kept constant across the different noise conditions. To ensure consistency in the inertia and viscoelasticity properties between the robot-robot, human-robot and human-human interactions, the experiment was conducted with an experimenter's relaxed right hand placed in one handle.

A robot was connected by a virtual spring to the other robotic partner that tracks the same target movement. The interactive torque (in Nm) was set to
\be \label{e:torque}
\tau(t) = \,k \,\left[\,q(t)-\tilde{q}(t)\,\right] \, ,
\ee
where $q$ and $\tilde{q}$ (in radian) denote the two robots' angular positions and the connection stiffness $k = 17.2\,\text{Nm/rad}$. 

The experiment consisted of 3\,(controllers)\,$\times$\,(8$\times$8 noise conditions) = 192 interactive trials. Each trial had a randomly determined controller \{SOIE, high impedance, low impedance\} and one of the noise conditions. Control performance was evaluated using the \textit{tracking error}
\be \label{e:error}
\left(\!\frac{1}{T} \! \int_0^T \!\!\! \left[q^*\!(t) - q(t) \right]^2 dt \!\right)^\frac{1}{2} \!\!,\, \quad T \equiv 10\,\text{s}
\ee
computed over each trial, and the tracking performance was evaluated using the sum of errors in the two robots. Furthermore, \textit{effort} over one trial was evaluated as
\be \label{e:effort}
\left(\!\frac{1}{T} \! \int_0^T \!\!\! \tau(t)^2 \, dt \!\right)^\frac{1}{2} \!\!,\quad T \equiv 10\,\text{s}\,.
\ee
To capture the measure of the strength of the desired signal relative to sensory noise, the signal-to-noise ratio is defined as the ratio of the power of a signal to the power of noise: 
\be
SNR = \,10 \log_{10} \!\left( \frac{P_{\text{signal}}}{P_\text{noise}} \right)
\ee 
where $P_{\mathrm{signal}}$ and $P_{\mathrm{noise}}$ are the average signal and noise power, respectively.

%%%%%%%%%%%%%%%%%%%%%%%%%%%%%%%%%%%%%%%%
\subsection{Human-robot experiment} \label{sect:human-robot protocol}
\subsubsection{Experiment protocol and procedure} 
The human-robot collaboration experiment was approved by the Joint Research Compliance Office at Imperial College London. Twelve participants (three female and nine male) without known sensorimotor impairments, aged 18\textendash44 years ($27.0$ $\pm$ $6.28$), were recruited. Their hand dominance was determined using the Edinburgh handedness inventory score \cite{oldfield1971assessment}, indicating that nine were right handed (with lateralisation quotient $\geq$ $60$), one was left handed, and one showed no clear dominance.  Each participant gave written informed consent prior to participation. 

Each participant was seated comfortably on a height-adjustable chair and held a handle of the Hi5 dual robotic interface with their dominant wrist while receiving visual feedback of the target movement on the front monitor. Medically certified surface electromyographic (EMG) was used to collect data from their two antagonist wrist muscles: the flexor carpi radialis (FCR) and extensor carpi radialis longus (ECRL) muscles. The EMG data was recorded at 100\,Hz.

The experiment consisted of three stages: calibration, solo trials and interaction trials. In calibration \cite{Takagi2018}, subjects were asked to flex or extend their wrist while the handle kept their wrist locked at zero degree, marked first as the subject’s most comfortable position. The subjects were asked to produce flexion and extension torques of 1, 2, 3 and 4 Nm for 2 seconds, first flexion then extension, with a rest period of 5 seconds between each activation to prevent fatigue. The measured EMG signal (in mV) was then mapped to a corresponding torque value (in Nm), so that the activity of each participant's flexor and extensor's can be compared and combined in the data analysis. After this calibration, the participants carried out 10 initial solo trials to learn the tracking task and the dynamics of the wrist interface. This was followed by 8 blocks of 10 interaction trials, where the robot was controlled by the developed SOIE control and a high impedance controller (HIC) randomly in the first 4 blocks and last 4 blocks. For each controller, the different noise conditions \{noisy(robot)-sharp(human): NS, SN, SS, NN\} were presented in a random order. The participants were informed when an experimental condition would be changed but not which condition would be encountered in the next trials. The tracking error defined in eq.\,(\ref{e:error}) was displayed at the end of each 20\,s trial to inform them the overall tracking performance of a trial, facilitating their adaptations to these noise conditions.   

After each trial, the target disappeared and the participants needed to place their respective cursor on the indicated starting position at the center of the screen. The next trial then started after a 5\,s rest period and a 3\,s countdown. The initialization of the next trial only started when the participant placed their wrist on the starting position, so that the participants could take a break at will in between trials, by keeping the cursor away from the center of the screen.

In \textit{interaction trials}, the partner's wrist was connected by a virtual spring with the same stiffness ($17.2$\,Nm/rad) as in the robot-robot experiment to a robotic partner with the interactive torque in eq.\,(\ref{e:torque}). The interaction trials were carried out under one of the two different visual feedback conditions on the human side which were defined as in the human-human experiment of \cite{Hendrik2022}. In the \textit{sharp human condition} the target was displayed as an 8\,mm diameter disk. In the \textit{noisy human condition} the target trajectory was displayed as a ``cloud'' of eight normally distributed dots around the target. The cloud of dots was defined by three parameters, randomly picked from independent Gaussian distributions: the vertical distance to the target position $\eta \in \mathcal{N}($0,\,[15\,mm]$^2 )$, the angular distance to the target position $\eta_q \in \mathcal{N}($0,\,[7.01$^\circ$]$^2)$, and the angular velocity $\eta_{\dot{q}} \in \mathcal{N}($0,\,[$10.01^\circ$/s]$^2)$. Each of the eight dots was sequentially replaced every 100\,ms. Two haptic noise conditions were presented for the robot. In \textit{noisy robot condition} there was biased Gaussian noise with mean $\delta = 7.01^\circ $ and standard deviation $\sigma = 0.05^\circ$ added to the reference target trajectory while the accurate target trajectory was given to the robot for tracking in \textit{sharp robot condition}.
\subsubsection{Evaluation of human behaviour} 
The torque of the flexor muscle was modelled from the envelope of the rectified and filtered EMG activity $u_f$ as in \cite{Hendrik2022}
\begin{equation}
	\begin{aligned}
		\tau_f(t) = \, a \, u_f(t) \, + \, b \,, \quad a, b > 0 \,,
	\end{aligned}
	\label{eq:regression}
\end{equation}
and similarly for the torque of the extensor muscle $\tau_e$. The human effort was then calculated as the averaged coactivation of the wrist flexor and extensor \cite{Franklin2008}
\be \label{e:humaneffort}
\!\frac{1}{T} \! \int_0^T \!\!\! \tau_{f}(t) + |\tau_{e}(t)| \,\, dt \, ,\quad T \equiv 20\,s \,.
\ee

To evaluate the overall cost of a human user in each trial when carrying out a tracking task, we used the weighted normalised tracking error and effort to make it comparable across participants. The tracking error was normalised as: 
\be \label{eq_normalisation}
e_n \equiv \frac{\overline{e} - \overline{e}_{min}}{\overline{e}_{max}-\overline{e}_{min}}\,
\ee
with $\overline{e}_{min}$ and $\overline{e}_{max}$ the minimum and maximum of the means of all trials of the specific participant. The normalised effort $\tau_n$ was calculated in the same way. In consideration of the error and effort balance, human cost was then computed as with the weights in \cite{Franklin2008}:
\be
J_h = 0.70\, e_n + 0.30\, \tau_n \,.
\ee

\subsubsection{Predicted target movement} 
Let $I$ be the robot's inertia, $L$ its control impedance, $\tilde{\eta}$ the target as sensed by the robot, and $\tau$ the interaction torque with the human partner, the robot's dynamics can be modelled as 
\be
I\ddot{\theta} = -L(\theta-\tilde{\eta})+\tau \equiv -L(\theta-\hat{\eta}) \,.
\ee
The robot can predict the target from their own sensing $\tilde{\eta}$ and the interaction torque $\tau$ with the human partner as
\be \label{e:prediction}
\hat{\eta}  \equiv \tilde{\eta} + \frac{\tau}{L} \,.
\ee 

\subsubsection{Statistical analysis} \label{sect: statistics}
Linear mixed effect (LME) analysis showed that after an initial learning effect, participants' error and effort stabilized over the last four trials of each block (non-significant slopes, $p > 0.05$). Therefore all subsequent analysis was conducted using the mean of the last 4 trials for each condition.

Shapiro-Wilk tests showed that tracking error and effort were not normally distributed in some noise conditions, so a Friedman test was used to examine the effect of the noise condition and controller. A three-way repeated-measures Aligned Rank Transform (ART) ANOVA with human visual noise, robot's sensory noise and controller as the factors was used in the analysis of the tracking error, effort and the overall cost. For post-hoc analysis after the ART ANOVA, paired Wilcoxon tests were conducted with the Holm-Bonferrroni correction.

\section{Supplementary figures}
\setcounter{figure}{0}
\renewcommand{\thefigure}{S\arabic{figure}}
\begin{figure}[thpb]
\includegraphics[width=\columnwidth]{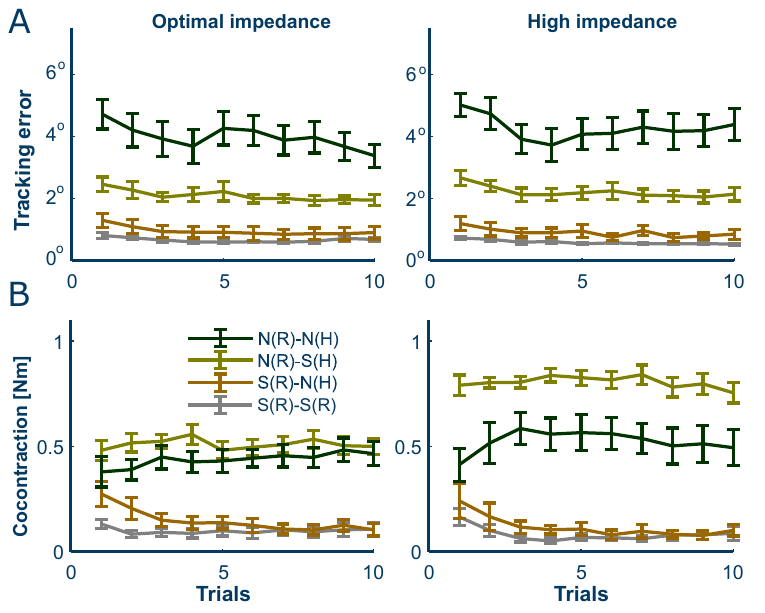}
\caption{Tracking errors and muscle cocontraction of SOIE (left) and high impedance controller (right) in the human-robot experiment. (A) Evolution of the tracking error over the 12 participants charted as a function of the trials, where the error bars represent one standard error on each side. When the robot agent has sharp sensory information, the tracking errors are similarly small for both controllers with little influence from human visual noise. However, when the robot agent has noisy sensory information, human visual noise has a larger impact on the tracking performance. Optimal impedance slightly improves the tracking in the NN condition as the error consistently decreases along trials while high impedance saturates has an increasing trend in after the first three trials. (B) Evolution of the normalized cocontraction as a function of trials, where the error bars represent one standard error. Human effort is reduced when the robot has noisy sensory information (especially in the NS condition), providing effective assistance to human users.}
\label{f:hr_model_tremg}
\end{figure}

\begin{figure}[thpb]
\includegraphics[width=\columnwidth]{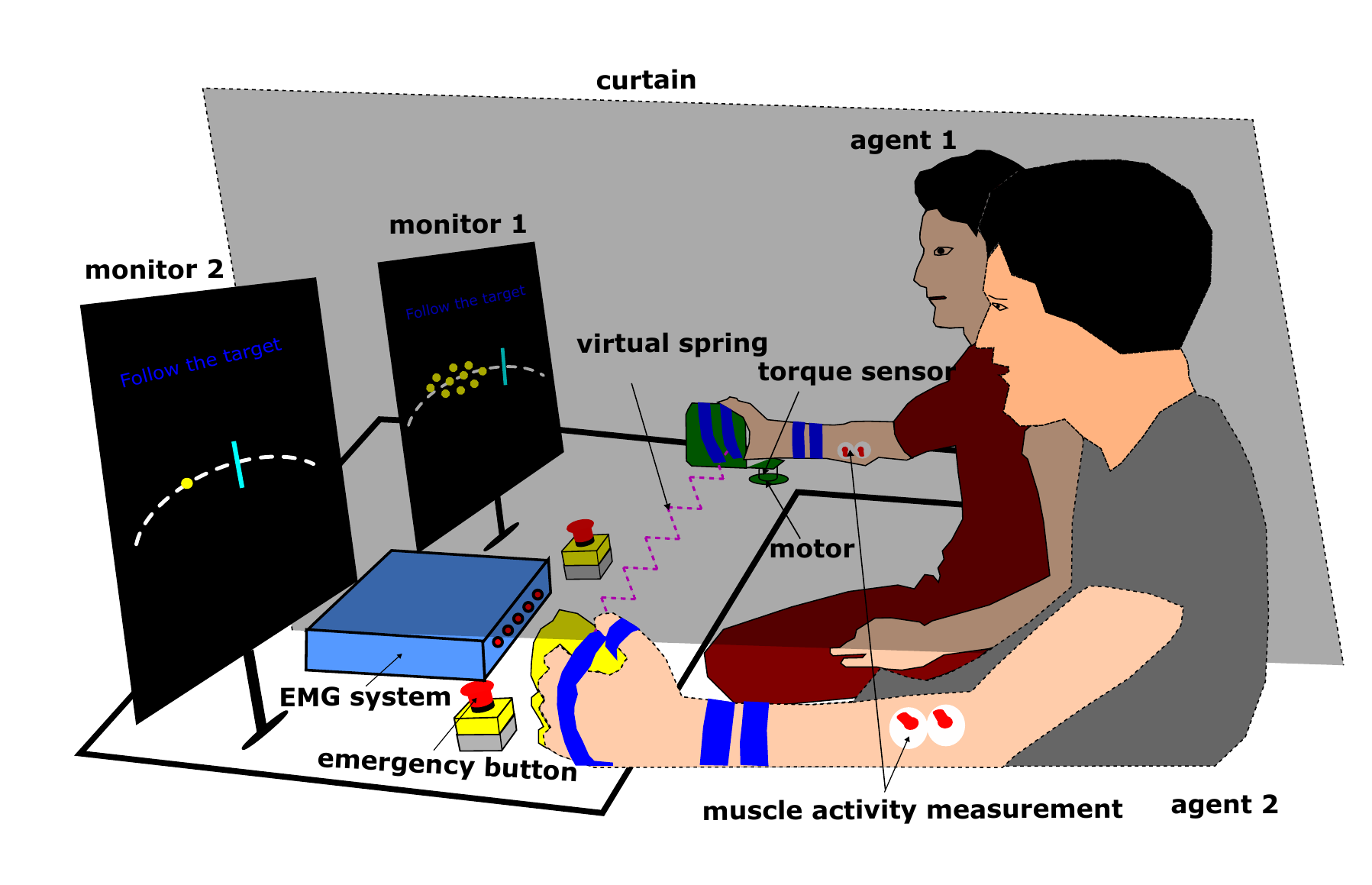}
\caption{xxx}
\label{f:hr_model_tremg}
\end{figure}

\section*{Acknowledgements}
We thank Vincent Hayward for suggestions on the algorithm validation, Nuria Peña Perez for discussions on statistical analysis, Gerolamo Carboni for discussions on the human-human experiment's results, and the participants for taking part in the experiments.

\subsection*{Funding}
\noindent EC grants PH-CODING (FETOPEN 829186)\\ EC grants CONBOTS (ICT 871803) \\
EC grants NIMA (FETOPEN 899626).

\subsection*{Author contributions}
\noindent Conceptualization: XC, JE, AT, EB.\\
Experiments: XC.\\
Data and statistical analysis: XC, JE, EB.\\
Computational modelling: XC, JE, BB, EB.\\
Funding acquisition: EB.\\
Visualisation: XC, JE, EB.\\
Writing: XC, JE, BB, AT, EB.

\bibliographystyle{IEEEtran}
\bibliography{IEEEabrv,autosam}

\end{document}